\documentclass{article}

\usepackage{arxiv}
\usepackage[utf8]{inputenc}
\usepackage[T1]{fontenc}
\usepackage{microtype}
\usepackage{amsfonts}
\usepackage{amsmath}
\usepackage{amssymb}
\usepackage{nicefrac}
\usepackage{graphicx}
\usepackage{booktabs}
\usepackage{tabularx}
\usepackage{multirow}
\usepackage{threeparttable}
\usepackage{url}
\usepackage{hyperref}
\usepackage{doi}
\usepackage{natbib}
\usepackage{cleveref}
\usepackage{lipsum}

\title{GeoAgentBench: A Dynamic Execution Benchmark for Tool-Augmented Agents in Spatial Analysis}

\date{} 					% Or removing it

\author{
	Bo Yu$^{1}$ \quad
	Cheng Yang$^{1}$ \quad
	Dongyang Hou$^{1}$\thanks{CONTACT: \texttt{houdongyang1986@csu.edu.cn}} \quad
	Chengfu Liu$^{1}$ \quad
	Jiayao Liu$^{1}$ \\
	\textbf{Chi Wang$^{1}$ \quad
	Zhiming Zhang$^{1}$ \quad
	Haifeng Li$^{1}$ \quad
	Wentao Yang$^{2}$} \\
	\\ % 空一行分隔作者与单位
	$^1$School of Geosciences and Info-Physics, Central South University, Changsha, China.\\
	$^2$School of Earth Sciences and Spatial Information Engineering, Hunan University of Science and Technology, Xiangtan, China.
}

\renewcommand{\shorttitle}{GeoAgentBench}

\hypersetup{
pdftitle={GeoAgentBench: A Dynamic Execution Benchmark for Tool-Augmented Agents in Spatial Analysis},
pdfsubject={cs.AI},
pdfauthor={Bo Yu, Cheng Yang, Dongyang Hou, Chengfu Liu, Jiayao Liu, Chi Wang, Zhiming Zhang, Haifeng Li, Wentao Yang},
pdfkeywords={GeoAI; Tool-Augmented Agents; Spatial Analysis; Benchmark},
}

\begin{document}
\maketitle

\begin{abstract}
	The integration of Large Language Models (LLMs) into Geographic Information Systems (GIS) marks a paradigm shift toward autonomous spatial analysis. However, evaluating these LLM-based agents remains challenging due to the complex, multi-step nature of geospatial workflows. Existing benchmarks primarily rely on static text or code matching, neglecting dynamic runtime feedback and the multimodal nature of spatial outputs. To address this gap, we introduce GeoAgentBench (GABench), a dynamic and interactive evaluation benchmark tailored for tool-augmented GIS agents. GABench provides a realistic execution sandbox integrating 117 atomic GIS tools, encompassing 53 typical spatial analysis tasks across 6 core GIS domains. Recognizing that precise parameter configuration is the primary determinant of execution success in dynamic GIS environments, we designed the Parameter Execution Accuracy (PEA) metric, which utilizes a “Last-Attempt Alignment” strategy to quantify the fidelity of implicit parameter inference. Complementing this, a Vision-Language Model (VLM) based verification is proposed to assess data-spatial accuracy and cartographic style adherence. Furthermore, to address the frequent task failures caused by parameter misalignments and runtime anomalies, we developed a novel agent architecture—Plan-and-React—that mimics expert cognitive workflows by decoupling global orchestration from step-wise reactive execution. Extensive experiments with seven representative LLMs demonstrate that the Plan-and-React paradigm significantly outperforms traditional frameworks, achieving the optimal balance between logical rigor and execution robustness, particularly in multi-step reasoning and error recovery. Our findings highlight current capability boundaries and establish a robust standard for assessing and advancing the next generation of autonomous GeoAI. The source code and dataset are available at: github.com/geox-lab/GABench.
\end{abstract}

% keywords can be removed

\section{Introduction}
Spatial analysis serves as one of the most fundamental tools in Geographic Information Science (GIScience), widely applied in critical scenarios such as urban planning, environmental monitoring, disaster assessment, and traffic management \citep{liao_spatiotemporal_2023, larkin_global_2023, pham_flood_2021, shahi_spatial_2023, shao_homogeneous_2024, guo_trimem_2025, cui_adversarial_2024}. However, geospatial analysis tasks are inherently challenging, frequently necessitating the integration of multi-source heterogeneous data and the execution of multi-step spatial computational workflows \citep{li_geospatial_2016, liakos_challenges_2022, wang_causal_2025, he_stdcformer_2025}. Driven by the rapid advancement in spatial data acquisition capabilities and technologies such as remote sensing and the Internet of Things (IoT), the complexity of modern geospatial analysis tasks has further escalated. Consequently, enhancing the automation and intelligence of spatial analysis workflows has emerged as a prominent research objective within the field of Geospatial Artificial Intelligence (GeoAI) \citep{janowicz_geoai_2020, peng_rethinking_2025}.

Unlike many isolated prediction tasks, real-world geospatial analysis typically manifests as complex workflows composed of diverse spatial operations. For instance, a comprehensive analytical pipeline may encompass multiple steps—such as data cleaning, coordinate reprojection, spatial overlay, spatial statistical modeling, and map visualization—which are often bound by strict logical dependencies. Traditional GeoAI research has predominantly focused on constructing specialized end-to-end models for single tasks, aiming to directly fit the input-output mapping through parameter optimization \citep{ronneberger_u-net_2015, kipf_semi-supervised_2017}. However, this paradigm overlooks the workflow heterogeneity that is pervasive in geospatial analysis tasks. Even when targeting the same geospatial objective, the analytical workflow is rarely static. The specific processing pathway is highly dependent on the characteristics of the data sources; for example, when input data consist of multi-source heterogeneous vector and raster formats, or operate across different spatial scales and coordinate reference systems (CRS)\citep{yao_estimating_2024}, the analytical pipeline must undergo dynamic adaptation and reconfiguration during stages such as data preprocessing, spatial operator selection, parameter configuration, and spatial statistical modeling. Constrained by their rigid architectures, traditional end-to-end models lack the orchestration capabilities required for such multi-step, non-linear analytical logic, rendering them ill-equipped to handle complex and variable real-world geographical scenarios. Consequently, in practical GIS operations, complex spatial analysis continues to rely heavily on manual planning and execution by domain experts using professional software. This reliance significantly hinders the democratization and automation of geospatial technologies\citep{li_autonomous_2023}, underscoring the urgent need for transition toward highly intelligent and autonomous geospatial systems\citep{li_earth_2017}.

In recent years, driven by the significant enhancement of Large Language Model (LLM) capabilities, constructing tool-augmented agents with LLMs as the central decision-making hub has emerged as a prominent research focus \citep{schick_toolformer_2023}. In contrast to traditional end-to-end models, these agents can comprehend user intent through natural language, decompose complex problems into a series of executable subtasks, and dynamically schedule external tools to implement computational logic. Currently, such agents have demonstrated substantial potential in domains including code generation, data analysis, and software automation \citep{zhang_codeagent_2024, zhang_data-copilot_nodate, xie_osworld_2024}. Building upon these successes, related research in the geospatial field has further corroborated that LLMs possess extensive geospatial knowledge and profound spatial reasoning capabilities \citep{roberts_gpt4geo_2023, mai_towards_2022}, while Vision-Language Models (VLMs) have also exhibited exceptional performance in the semantic understanding of remote sensing imagery \citep{lobry_rsvqa_2020, kuckreja_geochat_2024}. These advancements establish a robust cognitive foundation for the development of spatial agents. Consequently, introducing agents equipped with spatial reasoning and tool-invocation capabilities is regarded as a promising pathway to bridge the gap between general semantic reasoning and specialized spatial computation. Under this architecture, the agent leverages the LLM as its decision-making core for task decomposition and workflow planning, while precisely orchestrating external GIS tools for spatial computation. This transforms complex analytical pipelines—which traditionally relied on manual operation by domain experts—into natural language-driven automated processes, significantly lowering the barrier to entry for geospatial intelligence technologies \citep{huang_geoagent_2024}.

However, realizing this vision hinges on a critical prerequisite: the ability to systematically evaluate whether agents truly possess the capability to execute complex spatial analysis tasks. As the computational process entailing the highest cognitive complexity and the longest logical chains within GIS, spatial analysis imposes stringent demands on an agent’s planning proficiency, tool-use execution, and runtime error recovery capabilities. Although recent efforts in the academic community have yielded several relevant evaluation benchmarks—such as ToolBench\citep{qin_toolllm_2024} and API-Bank \citep{li_api-bank_2023} for general-domain API invocation, alongside GeoAnalystBench \citep{zhang_geoanalystbench_2025}, GeoBenchX \citep{krechetova_geobenchx_2025}, and GeoPlan-Bench \citep{li_designing_2025} tailored for geospatial tasks—these existing baselines still exhibit significant limitations when assessing authentic, complex spatial analysis workflows (as summarized in Table~\ref{tab:comparison}).

\begin{table}[htb]
    \caption{Comparison between GABench and existing general and geospatial-specific agent benchmarks.}
    \label{tab:comparison}
    \centering
    \begin{threeparttable}
    % --- 关键改进：定义一个居中的 X 列类型，确保占满全宽且内容居中 ---
    \newcolumntype{C}{>{\centering\arraybackslash}X} 
    
    \small % 12pt 下推荐使用 \small 或 \footnotesize
    \renewcommand{\arraystretch}{1.5} % 增加行高，让表格看起来更舒展
    
    % 使用 {l C C C C C}，其中 l 是左侧标题列，5个 C 均匀平分剩余宽度
    \begin{tabularx}{\textwidth}{l C C C C C}
        \toprule
        & \textbf{General Benchmarks\tnote{1}} & \textbf{GIS Code Gen\tnote{2}} & \textbf{GIS Planning Gen\tnote{3}} & \textbf{GeoPlan-Bench\citep{li_designing_2025}} & \textbf{GABench} \\ 
        \midrule
        \textbf{Domain Focus} & General API & Geospatial & Geospatial & Geospatial & Geospatial \\
        \textbf{Interaction} & Tool-Aug. & Code Gen & Text Plan & Tool-Aug. & Tool-Aug. \\
        \textbf{Environment} & API Calling & $\times$ & $\times$ & Mock Tools & Interactive Sandbox \\
        \textbf{Spatial Logic} & Missing & Moderate & Low & Low & High \\
        \textbf{Core Metrics} & API Acc. & Similarity & Trajectory & Trajectory & Trajectory \& VLM \\
        \textbf{Error Recovery} & Partial & Limited & None & None & Self-correction \\
        \textbf{Multimodal} & $\times$ & $\times$ & $\times$ & $\times$ & Dual Alignment \\
        \bottomrule
    \end{tabularx}
    
    \begin{tablenotes}
        \small
        \item[1] e.g., ToolBench\citep{qin_toolllm_2024}, API-Bank\citep{li_api-bank_2023}; 
        \item[2] e.g., GeoAnalystBench\citep{zhang_geoanalystbench_2025}, GeoCode-Bench\citep{hou_can_2025}; 
        \item[3] e.g., GeoBenchX\citep{krechetova_geobenchx_2025}.
    \end{tablenotes}
    \end{threeparttable}
\end{table}

Firstly, existing evaluations predominantly adopt non-interactive paradigms, severely lacking dynamic feedback loops within authentic execution environments. Specifically, current methodologies exhibit notable limitations in their evaluation mechanisms: (1) Planning-oriented workflow evaluation, such as GeoBenchX \citep{krechetova_geobenchx_2025}, which solely verifies the logical coherence of task plans at the textual level while neglecting their practical executability; (2) Surface-level measurement of code similarity, such as GeoAnalystBench \citep{zhang_geoanalystbench_2025}, which focuses on the lexical matching between model-generated scripts and expert reference codes, failing to assess the runtime efficacy of the code within genuine geospatial environments; and (3) Mocked validation of simulated invocations, such as GeoPlan-Bench \citep{li_designing_2025}, which relies on simulated environments to return mock tool execution results rather than executing them within actual GIS software infrastructures.

However, real-world spatial analysis environments are inherently fraught with uncertainties. Even logically sound plans may fail at runtime due to anomalies such as Coordinate Reference System (CRS) mismatches, invalid spatial topologies, or data format conflicts \citep{longley_geographic_2015}. In such instances, agents must rely on real-time feedback to perform error diagnosis and dynamic adjustments. Consequently, non-interactive evaluations not only fail to accurately gauge the agent's actual performance in complex geospatial tasks, but also struggle to capture its crucial capabilities in autonomous debugging and self-correction when confronting domain-specific runtime errors \citep{shinn_reflexion_2023}.

Secondly, existing benchmarks exhibit a pronounced deficiency in systematic coverage, primarily manifesting as an inadequate deconstruction of the complexities inherent in geospatial analysis. Currently, the majority of evaluation frameworks for geospatial tasks originate from the remote sensing domain \citep{shabbir_thinkgeo_2025, feng_earth-agent_2025}, where the incorporated GIS tasks frequently serve merely as auxiliary components. These tasks are predominantly confined to rudimentary operations such as area calculation, distance measurement, or basic spatial relationship assessments. Although GeoBenchX \citep{krechetova_geobenchx_2025} attempts to construct multi-step tasks tailored for commercial GIS practitioners—categorizing them by difficulty into four distinct tiers: "merge-visualize," "process-merge-visualize," "spatial operations (e.g., buffer and overlay analysis)," and "heatmap and contour generation"—the breadth of business scenarios it covers remains markedly limited. These tasks fail to adequately capture the highly intricate business requirements of real-world geospatial analysis. Instead, they are typically presented as combinations of isolated operators, lacking a profound simulation of complex spatial logical chains. Such oversimplification of analytical tasks renders it exceedingly difficult for existing benchmarks to systematically evaluate the comprehensive capabilities of agents when navigating authentic, long-chain geospatial analysis workflows.

Finally, existing evaluation methodologies predominantly oversimplify the assessment process into text-level comparisons, thereby neglecting the multimodal nature of spatial analysis deliverables. A comprehensive GIS analytical workflow typically yields not only textual outputs but also spatial data files (e.g., GeoJSON or TIFF) alongside ultimate cartographic representations. Nevertheless, the geometric correctness of such spatial datasets and the quality of their cartographic rendering are currently rarely integrated into a unified evaluation paradigm.

\begin{figure}[htb] % GSIS 建议使用 [htb] 或不加位置描述符
    \centering
    % 使用 \textwidth 确保图片宽度自适应单栏页面的边距
    \includegraphics[width=1.0\textwidth]{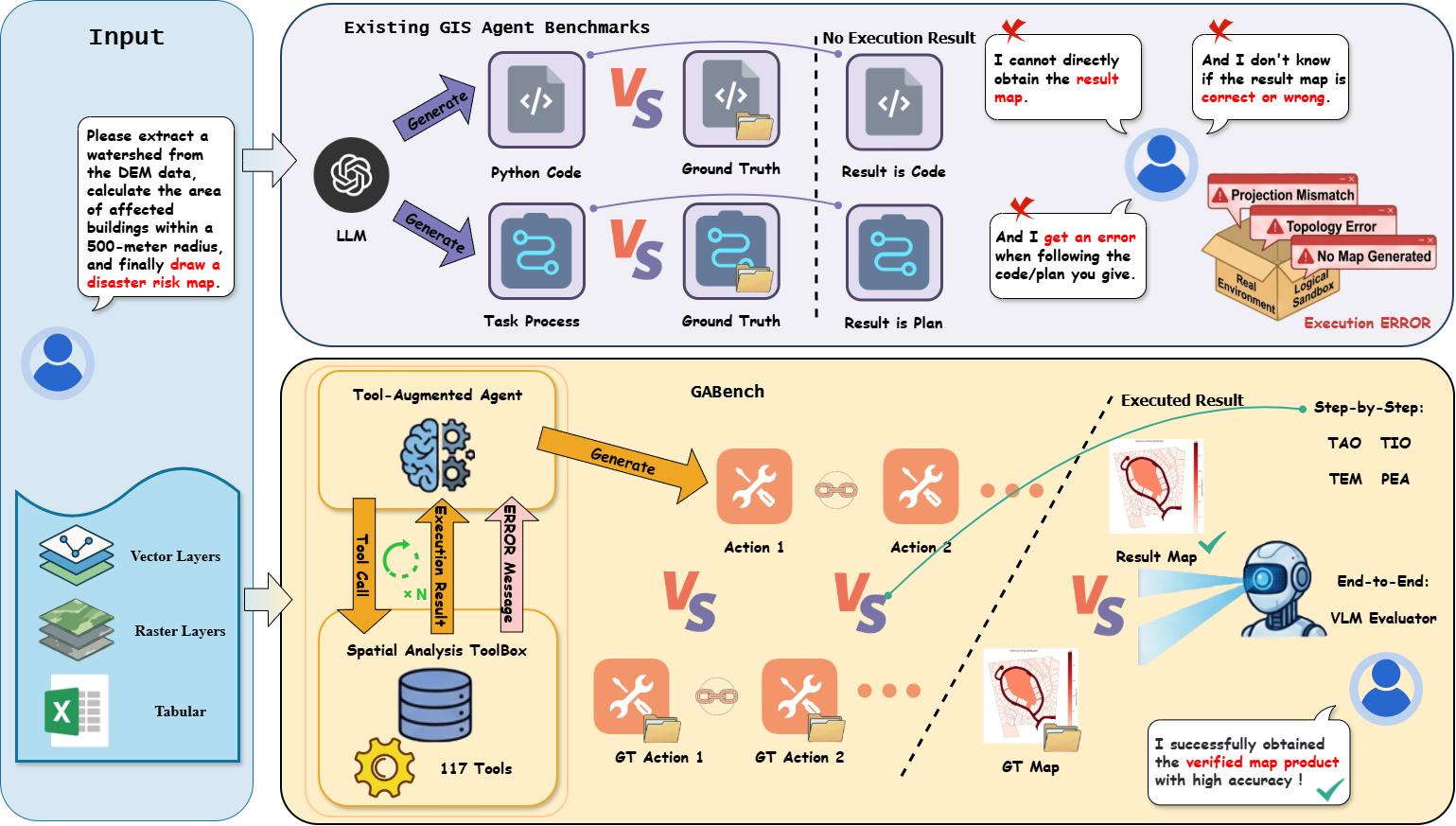} 
    \caption{Overview of the GeoAgentBench (GABench) framework compared with existing paradigms. The upper panel illustrates the limitations of traditional non-interactive benchmarks: LLMs generate code or plans that are evaluated via lexical similarity (Textual Ground Truth) but frequently encounter Execution Errors (e.g., projection mismatches, topology anomalies) in real GIS environments, leaving users with unverified and often unusable scripts. The lower panel showcases GABench’s dynamic, closed-loop execution paradigm. Our framework integrates a library of 117 atomic GIS tools within an interactive sandbox, enabling agents to perceive real-time Execution Feedback (Error Messages) and iteratively refine their trajectories. Evaluation is performed across two complementary tiers: (1) Step-by-Step metrics (TAO, TIO, TEM, and the innovative PEA) that verify the logical consistency of tool-invocation actions; and (2) End-to-End multimodal verification using a VLM Evaluator, which compares the final Result Map against the GT Map to ensure both data-spatial accuracy and cartographic quality, delivering a truly verified geospatial product.}
    \label{fig:teasor}
\end{figure}

To bridge the aforementioned research gaps, we introduce GeoAgentBench (GABench), an evaluation benchmark tailored for spatial analysis agents within dynamic and interactive environments. Diverging from conventional methodologies that rely on static textual matching, GABench is explicitly designed for tool-augmented agents, as illustrated in the comparison of execution paradigms in Fig~\ref{fig:teasor}. It constructs a closed-loop execution environment that integrates a comprehensive library of atomic GIS tools with authentic tool execution mechanisms. Within this benchmark, each test case comprises a natural language task description alongside multi-source spatial data. The agent is required to autonomously formulate analytical workflows based on user instructions, dynamically invoke and compose specialized GIS tools within the sandbox to execute spatial computations, and ultimately generate map visualizations as final outputs.

Our main contributions are as follows:

(1) We develop a dynamic and interactive evaluation benchmark (GABench) for tool-augmented GIS agents. By integrating 117 atomic tools and 53 representative tasks across 6 core GIS domains within a professional execution sandbox, GABench transcends traditional static text or code matching paradigms. This provides a systematic platform to assess an agent’s capacity for long-chain orchestration, implicit parameter inference, and execution-feedback-driven error recovery in complex real-world geospatial workflows.

(2) We develop a multi-tiered evaluation system that advances spatial agent assessment through the Parameter Execution Accuracy (PEA) metric and VLM-based verification. By integrating trajectory-level execution fidelity (via PEA) with end-to-end multimodal product validation (via VLMs), this system provides a more rigorous standard for quantifying both the precision of tool-level configurations and the cartographic quality of geospatial deliverables.

(3) We design a novel "Plan-and-React" agent architecture tailored for geospatial reasoning. Inspired by the cognitive workflows of human GIS professionals, we propose a specialized agent framework that decouples global workflow orchestration from local reactive execution. By anchoring dynamic "Thought-Action-Observation" loops within a pre-defined analytical blueprint, this architecture effectively mitigates the reasoning drift of pure ReAct and the execution rigidity of Plan-and-Solve, establishing a robust baseline for addressing highly heterogeneous and uncertain spatial tasks.

(4) We systematically analyze the capability boundaries of mainstream LLMs across four representative agent paradigms. Through extensive experiments under the Base Agent, ReAct \citep{yao_react_2023}, Plan-and-Solve \citep{wang_plan-and-solve_2023}, and Plan-and-React frameworks, this work reveals significant disparities in multi-step reasoning and error recovery, establishing a benchmark for the next generation of autonomous GeoAI systems.

\section{Related Work}
\subsection{Geospatial Foundation Models}
In recent years, general-purpose foundation models, particularly Large Language Models (LLMs), have achieved breakthroughs in natural language processing and complex logical reasoning. Furthermore, Vision-Language Models (VLMs) have extended single-text modalities to multimodal understanding, endowing models with robust visual perception and cross-modal interaction capabilities. Leveraging massive training datasets, these general-purpose models demonstrate exceptional generalization and zero-shot/few-shot learning capabilities, laying the groundwork for the paradigm shift of geospatial intelligence from domain-specific small models to general-purpose intelligent systems \cite{huang_role_2026}.

However, due to the inherent high heterogeneity, spatial dependency, and complex topological rules of geospatial data, the direct application of general-purpose models often fails to fully capture the deep semantics of the geospatial domain \citep{mai_towards_2022, ji_foundation_2025}. Consequently, researchers have recently begun actively exploring foundation models customized for the geospatial field \citep{janowicz_geofm_2025}. In language modeling, geospatial foundation models such as K2 \citep{deng_k2_2024} have significantly enhanced their understanding of geoscientific knowledge through continual pre-training and instruction tuning on corpora containing millions of earth science documents. Additionally, studies like GeoLLM \citep{manvi_geollm_2024} have demonstrated the feasibility of directly extracting and augmenting geospatial knowledge from LLMs by injecting OpenStreetMap vector network and coordinate information into prompts. In the visual and multimodal domains, Vision-Language Models for remote sensing imagery have also emerged. By aligning remote sensing imagery with natural language descriptions, these models have exhibited outstanding performance in perception-oriented tasks such as remote sensing Visual Question Answering (VQA), scene classification, and object localization \citep{li_ddfav_2025, an_choice_2024}.Furthermore, AllSpark \citep{shao_allspark_2025} proposed a unified framework supporting ten heterogeneous spatio-temporal modalities, further expanding the collaborative understanding capabilities of geospatial foundation models for multi-source data.

Despite significant progress in professional knowledge acquisition and multimodal perception, geospatial foundation models remain limited in their ability to support complex spatial analysis. Firstly, current geospatial foundation models primarily focus on factual question answering, text summarization, or basic geographic entity recognition, leaving their spatial reasoning and tool-invocation capabilities largely underexplored. Authentic, complex spatial analysis relies heavily on multi-step, integrated calls to professional GIS operators and external software (e.g., GDAL, GeoPandas, or spatial databases). However, most open-source geospatial foundation models lack high-quality training data on GIS tool-invocation logic chains during the instruction-tuning phase, rendering them incapable of handling complex operations such as parameter configuration, topological error detection, and runtime self-correction.

\subsection{Geospatial Intelligent Agents}
In recent years, autonomous agents based on Large Language Models (LLMs) have emerged as a core research focus in the field of artificial intelligence. Unlike traditional LLMs that function merely as static text generators, these agents integrate modules such as memory mechanisms, task planning, and external tool invocation to achieve dynamic perception and interaction with their environments \citep{liu_survey_2025}.

Inspired by the development of general-purpose agents, scholars in geospatial science are actively exploring the construction of specialized agents for GIS and complex remote sensing tasks. In the GIS domain, researchers have proposed theoretical frameworks for "Autonomous GIS," aimed at leveraging the language understanding and code generation capabilities of LLMs to automate the full lifecycle of tasks, from spatial data discovery and collection to analysis and visualization \citep{li_autonomous_2023}. Following this paradigm, a series of native geospatial agent systems have emerged. For example, LLM-Find, an agent framework focused on geographic data retrieval, can autonomously search for and download formatted geospatial data from predefined open-source interfaces by executing and debugging code based on natural language instructions \citep{ning_autonomous_2025}. Similarly, the GIS Copilot system embeds LLMs into open-source GIS platforms (e.g., QGIS), enabling non-expert users to generate and execute spatial analysis code through dialogue \citep{akinboyewa_gis_2025}. Furthermore, multi-agent systems such as ShapefileGPT have achieved efficient processing loops for specific vector data types (e.g., Shapefiles) by decoupling complex task planning from specific tool invocation \citep{lin_shapefilegpt_2025}.

In the remote sensing domain, agent technology is widely employed to reduce the barriers and operational complexity associated with processing multimodal observational data. For instance, addressing the challenge of selecting from a vast array of remote sensing foundation models, the recently proposed REMSA agent can autonomously retrieve, match, and recommend the most suitable model from a meta-database based on user natural language requirements \citep{chen_remsa_2025}. Meanwhile, multi-agent systems like GeoLLM-Squad have established coordination and allocation mechanisms to distribute complex remote sensing analysis workflows to specialized sub-agents for retrieval, analysis, and visualization, significantly improving processing efficiency for large-scale remote sensing imagery \citep{lee_multi-agent_2025}.

Despite the immense potential demonstrated by geospatial agents in automated map production, data retrieval, and fundamental spatial analysis, current systems exhibit significant limitations when handling real-world, highly complex spatial workflows. On one hand, existing spatial agents rely heavily on the zero-shot code generation capabilities of general-purpose LLMs, yet lack mechanisms for sensing and self-correcting spatial analysis-specific errors. When faced with runtime errors—such as coordinate system mismatches, invalid spatial topologies, or misconfigured spatial parameters—these agents often struggle to diagnose the issues or resume execution autonomously. On the other hand, while current spatial agents perform adequately in tasks involving single operators or explicit step-by-step guidance, their ability to decompose tasks and dynamically orchestrate long-chain, multi-step spatial workflows with unknown steps and complex dependencies remains weak \citep{akinboyewa_gis_2025}. This further underscores the urgent need for systematic evaluation of the comprehensive capability boundaries of agents in complex geospatial scenarios.

\subsection{Benchmarking for Geospatial Intelligent Agents}
With the advancement of tool-learning capabilities in large models, mature benchmarks such as ToolBench \citep{qin_toolllm_2024} and API-Bank \citep{li_api-bank_2023} have been established in general-purpose domains. Although these benchmarks provide a wide array of API-calling scenarios, they primarily treat tools as stateless general functions, making it difficult to capture the inherent spatial-semantic constraints of geospatial analysis, such as coordinate system transformation logic between multi-source data or validity checks for geometric topology.

In the vertical exploration of the geospatial field, several agent-based and evaluation studies targeting remote sensing imagery have recently emerged (e.g., ThinkGeo \citep{shabbir_thinkgeo_2025}, Earth-Agent \citep{feng_earth-agent_2025}). However, these works primarily focus on the visual perception of raster data and image-based semantic question answering—essentially, information extraction from imagery. Real-world GIS applications require not only the ability to recognize geographic objects but also rely on the dynamic invocation and logical reasoning of complex vector data, topological relationships, and multi-step spatial analysis toolchains.

In the domains of code generation and data analysis, works such as DS-1000 \citep{lai_ds-1000_2023} have demonstrated that execution-based evaluation outperforms static text evaluation. Nevertheless, regarding specialized evaluations for the geospatial domain, existing research is generally constrained by static evaluation paradigms, lacking authentic interaction and execution. Specifically, whether generating task plans in natural language (e.g., GeoBenchX \citep{krechetova_geobenchx_2025} and GeoPlan-Bench \citep{li_designing_2025}) or Python scripts (e.g., GeoAnalystBench \citep{zhang_geoanalystbench_2025} and GeoCode-Bench\citep{hou_can_2025}), current benchmarks largely score by calculating text similarity or code-matching degrees between generated content and static ground truths. This assessment approach, detached from a real execution environment, suffers from a severe lack of situational grounding: a piece of code or a plan that is syntactically similar to the ground truth may still suffer runtime crashes when processing highly heterogeneous spatial data, due to factors like parameter threshold sensitivity, empty geometries, or file-locking issues. Due to the absence of closed-loop feedback from a real execution environment, existing benchmarks fail to quantify an agent’s robustness and self-correction capability when facing runtime errors.

Furthermore, existing evaluation systems lack focus on the quality of cartographic visualization. The final output of geographic analysis is often a map. While attempts have been made in the general chart domain (e.g., MatPlotBench \citep{yang_matplotagent_2024}), geographic maps involve complex projections and semiotics. Existing metrics remain primarily based on text matching and neglect the assessment of map aesthetics and the accuracy of symbolic representation, leading to an incomplete evaluation of an agent’s end-to-end capabilities.

To address these issues, the GABench proposed in this paper constructs a closed-loop execution environment integrated with a native GIS tool library. By incorporating a multi-dimensional task system, a dynamic evaluation mechanism based on runtime feedback, and the innovative introduction of Vision-Language Models (VLMs) for end-to-end multimodal product verification, this benchmark aims to establish a novel evaluation standard that is comprehensively aligned with the logic and complexity of real-world GIS applications.

\section{Benchmark Design}
\begin{figure}[htb] % GSIS 建议使用 [htb] 或 [t]
    \centering
    \includegraphics[width=\textwidth]{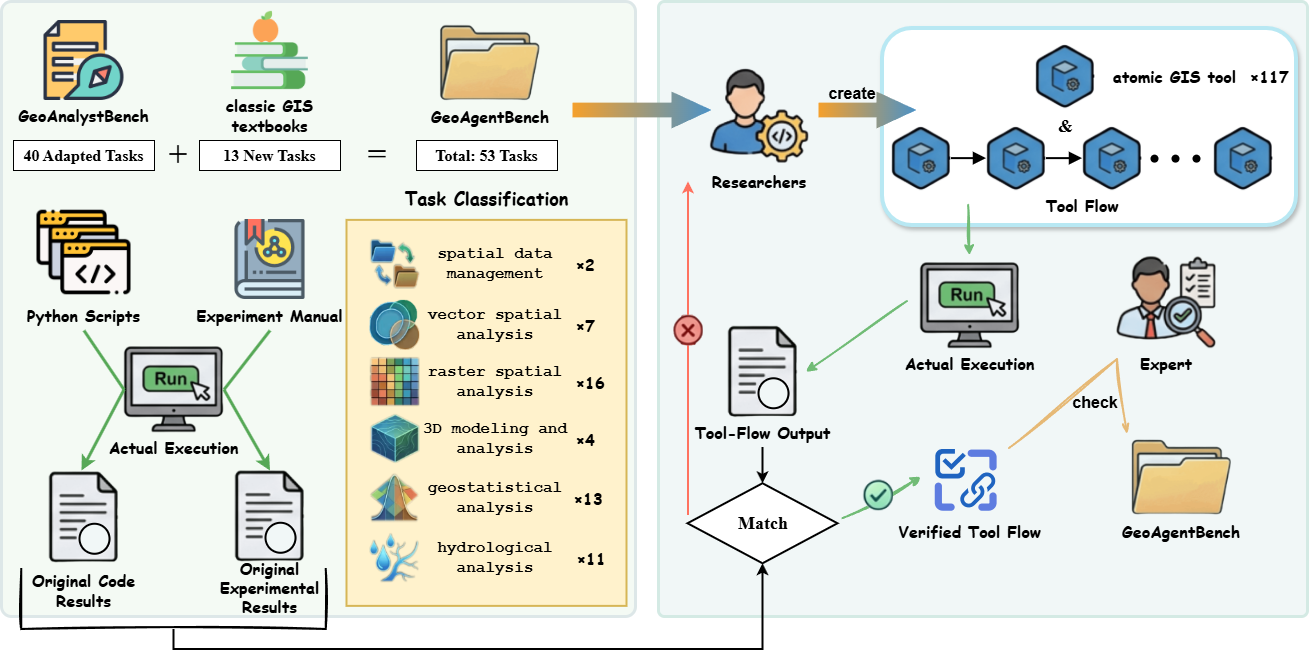} % 确保占满单栏宽度
    \caption{Overview of the GABench dataset construction and verification workflow. The left panel outlines task sourcing (53 total tasks) and the hierarchical taxonomy across six core GIS domains. The right panel depicts the iterative structural refactoring and verification process: researchers first develop 117 atomic GIS tools and re-engineer the original analytical logic into standardized, modular tool-flows. A closed-loop consistency check is then performed by matching tool-flow outputs against original code and experimental results; any discrepancies trigger iterative refinement by the researchers (indicated by the red arrow). Finally, an expert review ensures logical integrity and functional redundancy screening, certifying the verified tool-flows as the physical ground truth for the GABench dataset.}
    \label{fig:pipeline}
\end{figure}

The design of GABench follows a modular and integrated architecture aimed at bridging the gap between high-level reasoning and physical geospatial computation. As illustrated in Fig~\ref{fig:pipeline}, the benchmark comprises a hierarchical task system grounded in professional GIS domains and a dynamic execution sandbox powered by a library of atomic GIS tools. This section details the systematic construction of GABench, from its task taxonomy and expert-led refactoring to its standardized metadata architecture.

\subsection{Task Categories and Sources}
To ensure that the evaluation benchmark comprehensively covers the core capabilities required for complex geospatial analysis, this study constructed a hierarchical task taxonomy grounded in classic GIS literature and textbooks \citep{_arcgis2_2012} (as illustrated in Fig~\ref{fig:pipeline}). This taxonomy not only encompasses both vector and raster data models but also achieves a progressive transition in logical depth—from basic geometric operations to complex process simulations. It is categorized into six core domains: (1) spatial data management; (2) vector spatial analysis; (3) raster spatial analysis; (4) 3D modeling and analysis; (5) geostatistical analysis; and (6) hydrological analysis. This hierarchical architecture establishes a rigorous baseline for systematically evaluating an agent’s long-chain reasoning and multi-step tool orchestration capabilities.

The construction of the task set began with a rigorous screening of GeoAnalystBench \citep{zhang_geoanalystbench_2025}. We evaluated its 50 cases and excluded 10 tasks that relied on closed-source data or proprietary ArcGIS formats, ensuring compatibility with open geospatial computing ecosystems. However, a deeper analysis revealed that the remaining 40 cases could not be directly utilized for autonomous tool orchestration due to two critical limitations: (1) Coarse Logical Granularity: The original human-designed workflows were organized into high-level logical abstractions—such as "load dataset," "interpolate," or "merge"—rather than fine-grained, executable operations. This lack of atomic granularity makes it difficult for agents to dynamically schedule sub-tasks; and (2) Monolithic Scripting: The accompanying Python code consisted primarily of task-specific monolithic scripts, which lacked the modularity required for decomposition into reusable atomic tools. Consequently, rather than utilizing the original source code, we opted for a complete reconstruction of the analytical logic—a process of structural refactoring detailed in the following section. To better support this transition and enable end-to-end verification, we reshaped the task descriptions by modifying instructions to mandate cartographic outputs and standardizing all file access paths.

To further address coverage deficiencies in the original benchmark—particularly within complex hydrological analysis scenarios—we introduced 13 high-difficulty tasks systematically sourced from classic GIS textbooks \citep{_arcgis2_2012}. This targeted expansion ensures that GABench provides comprehensive coverage across all six core GIS domains, ultimately resulting in a suite of 53 representative tasks with explicit geographical significance. 

\subsection{Stepwise Tool Chains and Sandboxed Environment for Tasks}
To overcome the limitations of monolithic scripts and coarse logical abstractions, we performed a profound structural refactoring of the task set. The significant leap in granularity and logical rigor is best illustrated through a typical Urban Heat Island analysis task. In GeoAnalystBench, the workflow is defined by seven high-level semantic steps: (1) Load dataset, (2) Interpolate, (3) Filter, (4) Merge, (5) Average, (6) Highlight, and (7) Visualization. While these steps capture the semantic intent, they suffer from critical logical gaps and lack the execution-level details necessary for an autonomous agent. For instance, the transition from "Interpolate" (which generates a continuous raster surface) to "Average" (which requires polygon-based aggregation) involves an implicit spatial data model conflict that traditional benchmarks overlook.

In contrast, GABench refactors this task into a precise, atomic Tool Flow composed of 117 standardized GIS tools. As shown in our refactored workflow, the "Interpolate" step is explicitly handled by the \textit{ordinary\_kriging} tool, requiring defined grid bounds, resolution parameters (e.g., \textit{nx=100}, \textit{ny=100}), and specific variogram models. To resolve the data model conflict, we introduce a \textit{zonal\_statistics} operation as a computational bridge to aggregate raster heat values into the CensusBlock vector layer. Furthermore, the vague instruction to "Highlight" is translated into a rigorous two-step sequence: first, a \textit{filter\_features\_by\_expression} tool performs a precise spatial query; then, a \textit{create\_multilayer\_map} tool handles the complex cartographic logic by stacking the base heat distribution with high-risk highlight layers using specific visual arguments (e.g., OrRd color maps and alpha transparency).

This atomic design ensures that every step corresponds to a precise geospatial operation, providing a rigorous and executable baseline. These tools were designed following the principles of universality, non-specificity, and high reusability. To ensure logical rigor, an expert-led review mechanism was introduced to conduct functional redundancy screening and verify the optimal tool-flow for each task. A consistency verification closed-loop was established to validate these trajectories. We used the execution deliverables—generated from original scripts \citep{zhang_geoanalystbench_2025} or classic GIS experiments \citep{_arcgis2_2012}-as the reference. We mandated that the outputs of our refactored workflows remain strictly consistent with these original results at both the semantic and data levels (e.g., matching geometric topologies and attribute precision). These validated outputs constitute the Verified Physical Ground Truth of GABench, facilitating a shift from surface-level code matching to autonomous tool orchestration.

To support the automated execution of these workflows, GABench provides a lightweight, interactive runtime sandbox leveraging a native Python open-source geospatial stack (including GeoPandas, Rasterio, and Shapely). This environment serves as a bridge between the agent and physical data through three core mechanisms. First, the sandbox allocates a unique contextual workspace for each task, where all intermediate data streams and outcomes are persisted in real-time via a persistent state management system. Second, we designed an isolated denoising feedback mechanism to mitigate hallucinations and parameter drifts. This mechanism intercepts complex Python tracebacks and distills them into semantically clear error messages before presenting them to the agent. Third, the environment enforces strict file-write policies to simulate realistic GIS resource conflicts, such as file-locking exceptions, which compel agents to perform autonomous error diagnosis and recovery within a genuine, feedback-driven execution environment.

\subsection{GABench Description}
To ensure reproducibility and systematic execution, each task in GABench is organized into a standardized metadata schema. As detailed in Table \ref{tab:metadata}, this schema serves as a bridge between high-level human instructions and machine-executable toolflows, encompassing several core fields. Each task is identified by a unique ID and categorized into a specific Domain (e.g., Vector Spatial Analysis). The Task Description provides the natural language instruction defining the analysis objective, while the Data Description offers standardized file paths and metadata for multi-source input datasets. To support end-to-end evaluation, the Drawing Style specifies cartographic requirements such as color ramps and layer order. Regarding execution logic, the Toolchain Length denotes the total number of atomic tool invocations in the gold-standard workflow, and the Toolchain JSON provides the structured sequence of tool calls and parameter configurations. Finally, the Result field identifies the filename of the generated map product for visual verification, while Layers describes the specific data layers integrated into the final rendering.

\begin{table}[htb] % 使用模板推荐的 [htb]
    \centering
    \caption{Description of the metadata schema for GABench tasks.} % 改为标准句式大小写
    \label{tab:metadata}
    \begin{threeparttable}
    \small % 12pt 下推荐使用 \small，如果内容还是多可以用 \footnotesize
    \renewcommand{\arraystretch}{1.3} % 1.3-1.4 比较合适
    \setlength{\tabcolsep}{6pt}      % 恢复到标准的 6pt 间距，视觉上更稳重

    % 使用 tabularx 确保占满整个页面宽度 (\textwidth)
    \begin{tabularx}{\textwidth}{@{}l X@{}}
        \toprule
        \textbf{Field} & \textbf{Description} \\
        \midrule
        ID & Unique identifier for each spatial analysis task. \\
        Domain & The specific GIS domain the task belongs to (e.g., Vector Spatial Analysis, Hydrological Analysis). \\
        Task Description & Natural language instruction defining the analysis objective and requirements. \\
        Data Description & Standardized file paths and metadata for the multi-source input datasets. \\
        Drawing Style & Specific cartographic requirements, including color ramps, layer order etc. \\
        Toolchain Length & The total number of atomic GIS tool invocations in the gold-standard workflow. \\
        Toolchain JSON & The sequence of tool calls and parameter configurations in a structured format. \\
        Result & The filename of the final generated map product used for visual verification. \\
        Layers & The specific data layers integrated into the final map rendering. \\
        \bottomrule
    \end{tabularx}
    \end{threeparttable}
\end{table}

Statistical analysis of the final 53 tasks highlights the professional depth and complexity inherent in GABench. The overall task chains involve an average of 6.7 tool invocations per task, with the Toolchain Length peaking at 17 steps. Furthermore, each task requires an average of 2.06 input layers, necessitating complex data-joining and overlay operations. This high density of tool-data interactions is specifically designed to evaluate agents across core dimensions including spatial commonsense comprehension, long-chain tool orchestration, and implicit parameter inference. Compared to previous benchmarks that focus on single-step operators, the multi-step nature of GABench better reflects the realistic complexity of professional GIS workflows.

To ensure high consistency and environment isolation during evaluation, the complete dataset and execution framework are constructed using modern engineering standards. We utilize the uv dependency manager to enforce strict version locking for the geospatial computational stack, including specific versions of GeoPandas and Rasterio. This standardized configuration effectively eliminates system-level interference caused by dependency conflicts, ensuring that diagnostic feedback and execution outcomes remain consistent across different experimental platforms. By providing this transparent and stable evaluation suite, GABench establishes a robust foundation for identifying the capability boundaries of autonomous spatial agents.

\section{Multi-Tiered Evaluation Metrics: Advancing Assessment with PEA and VLM}
To comprehensively quantify the performance of tool-augmented GIS agents within the dynamic sandbox, we propose a multi-tiered evaluation system that transcends traditional static text-matching paradigms. While building upon standard trajectory metrics such as TAO, TIO, and TEM, our system is anchored by two major technical advancements: the Parameter Execution Accuracy (PEA) metric and a multimodal VLM-based verification mechanism. Specifically, we assess agents across three critical dimensions: (1) Step-by-Step trajectory coherence, featuring PEA to precisely measure implicit parameter inference; (2) End-to-End product quality, leveraging Vision-Language Models (VLMs) as automated evaluators for data-spatial accuracy and cartographic style; and (3) Operational Efficacy to measure resource utilization. Together, these metrics provide a rigorous and objective standard for characterizing the capability boundaries of autonomous agents in handling real-world GIS complexities.

\subsection{Trajectory-level Evaluation and the PEA Metric}
To rigorously assess an agent’s performance in the logical orchestration of geospatial workflows, we implement a trajectory-level evaluation system. Following the trajectory assessment principles established in Earth-Agent \citep{feng_earth-agent_2025}, we employ TAO (Tools-Any-Order), TIO (Tools-In-Order), and TEM (Tools-Exact-Match) metrics to quantify the structural coherence of tool-invocation sequences. While these metrics effectively measure the accuracy of the planned path, they lack the granularity required to evaluate the precision of individual tool configurations. To bridge this gap, we introduce the Parameter Execution Accuracy (PEA) metric to capture the efficacy of an agent's tool-level parameterization. Unlike standard path-centric metrics, PEA is specifically designed to isolate the validity of the agent’s final successful invocation from intermediate trial-and-error logs. By utilizing a "Last-Attempt Alignment" strategy, it provides a more precise quantification of the agent’s implicit parameter inference capabilities within complex and highly uncertain geospatial tasks.

Tools-Any-Order (TAO): This metric aims to quantify an agent's capability in identifying and retrieving the necessary set of atomic spatial tools. It focuses on evaluating whether the agent can accurately pinpoint the core set of tool operators required to solve complex geospatial analysis tasks, independent of the order of invocation. We denote the set of tools retrieved by the agent during execution as $\mathcal{T}_{pred}$, and the set of tools in the Ground Truth as $\mathcal{T}_{gt}$. To balance the precision ($P$) and recall ($R$) of the predicted toolset, we employ the $F_{1}-Score$ as the comprehensive evaluation metric, which is defined as follows:

\begin{equation}
\label{eq1}
\begin{gathered}
F_{1}-Score=\frac{2\cdot P\cdot R}{P+R}, \\
\quad\mathrm{where}\quad P=\frac{|\mathcal{T}_{pred}\cap\mathcal{T}_{gt}|}{|\mathcal{T}_{pred}|},R=\frac{|\mathcal{T}_{pred}\cap\mathcal{T}_{gt}|}{|\mathcal{T}_{gt}|}
\end{gathered}
\end{equation}

Geospatial analysis tasks inherently possess strict logical dependencies (e.g., coordinate reprojection must precede area calculation). To accurately assess the logical completeness of the analytical workflow, this study introduces the following two dimensions for quantitative measurement:

Tools-In-Order (TIO): This metric aims to evaluate an agent's grasp of the sequential order of tool invocations within a spatial analysis workflow. Inspired by the concept of the Longest Common Subsequence (LCS), we calculate the proportion of standard tools that maintain their correct relative order within the predicted trajectory compared to the total number of steps in the standard workflow. This provides an objective reflection of the structural correctness of the workflow's topological logic. The metric is highly robust against non-destructive intermediate steps (e.g., data validation) inserted by the agent. The formula is defined as follows:

\begin{equation}
\label{eq1}
TIO=\frac{|\mathrm{LCS}(\mathbf{T}_{pred},\mathbf{T}_{gt})|}{|\mathbf{T}_{gt}|}
\end{equation}

\noindent where $\mathbf{T}_{pred}$ and $\mathbf{T}_{gt}$ denote the predicted and ground-truth tool-invocation sequences, respectively; $\mathrm{LCS}(\cdot)$ denotes the Longest Common Subsequence of the two sequences.

Tool-Exact-Match (TEM): This metric adopts a strict prefix-matching principle, aiming to measure the agent's precise adherence to the Standard Operating Procedure (SOP). It calculates the proportion of the tool-invocation sequence that remains perfectly identical to the ground truth from the very beginning of the analytical process. This metric provides an in-depth characterization of the accuracy of the analytical path, particularly under conditions of long-chain dependencies. The formula is defined as follows:

\begin{equation}
\label{eq1}
TEM = \frac{|\mathrm{LCP}(\mathbf{T}_{pred}, \mathbf{T}_{gt})|}{|\mathbf{T}_{gt}|}
\end{equation}

\noindent where $\mathbf{T}_{pred}$ and $\mathbf{T}_{gt}$ denote the predicted and ground-truth tool-invocation sequences, respectively; $\mathrm{LCP}(\cdot)$ denotes the Longest Common Prefix, representing the longest subsequence where the two sequences match continuously starting from the first element.

Parameter Execution Accuracy (PEA): This metric is specifically designed to quantify the precision of parameter configuration and the actual execution efficacy of an agent within critical workflows. Given that agents often exhibit "trial-and-error" behaviors guided by environmental feedback in complex tasks, traditional sequential comparison methods are prone to misjudgment due to intermediate failed attempts. To decouple true execution performance from trivial interaction logs, we innovatively propose a dual-stage computational paradigm: Backward Alignment and Forward Evaluation. First, in the Backward Alignment stage, we employ a reverse-retrieval strategy to align the steps of the ground truth with the agent’s final invocation of the corresponding tool at each logical node—a mechanism we term "Last-Attempt Alignment." The core rationale is that in long-chain workflows, only the final operation, after an agent's self-reflection and correction, represents the critical variable determining the actual outcome of that step. Subsequently, in the Forward Evaluation stage, we introduce a Dynamic Variable Mapping mechanism, which accounts for the inherent randomness in generated intermediate filenames while ensuring that the mapping remains strictly consistent within subsequent topological inputs. More crucially, to mitigate the risk of parameter "hallucination," we incorporate a Physical State Check: for key parameters involving file paths, the system verifies the physical existence of the file within the genuine output sandbox directory. Concurrently, for specific tools such as those involved in visualization, we strategically relax the inspection of non-deterministic stylistic parameters (e.g., titles) to accommodate the inherent stylistic diversity in outputs generated by different large models.The formula for PEA is defined as follows:

\begin{equation}
\label{eq1}
\begin{aligned}
PEA &= \frac{1}{N}\sum_{i=1}^{N} \mathbb{I} \left( \text{Tool}_{last}^{(i)} = \text{Tool}_{gt}^{(i)} \right. \\
&\quad \left. \wedge \, \text{Params}_{last}^{(i)} \cong_{\mathcal{M},\mathcal{E}} \text{Params}_{gt}^{(i)} \right)
\end{aligned}
\end{equation}

\noindent where $N$ represents the total number of steps in the ground-truth sequence; ${Tool}_{last}^{(i)}$ denotes the agent's final tool instance within the backward retrieval window corresponding to the $i$-th step of the ground truth; and $\cong_{\mathcal{M},\mathcal{E}}$ signifies the semantic equivalence assessment of parameters, subjected to the dual constraints of the dynamic variable mapping $\mathcal{M}$ and the physical file existence verification $\mathcal{E}$ within the sandbox. This metric effectively isolates the agent's final execution efficacy from redundant trial-and-error interactions. It not only precisely quantifies the agent's implicit parameter inference capabilities but also establishes a scientific baseline for assessing the real-world reliability of large language models, corroborated by execution feedback from a genuine sandbox environment.

\subsection{A New VLM-based End-to-End Metric}
While trajectory-level metrics quantify the logic of tool-invocation sequences, they remain insufficient for evaluating the ultimate efficacy of the geospatial deliverables. Since complex spatial analysis workflows typically culminate in visual map products, we introduce a multimodal automated verification mechanism powered by Vision-Language Models (VLMs). This approach allows for a rigorous, end-to-end assessment of both the data-spatial accuracy and the cartographic style of the generated maps. By utilizing VLMs as objective judges, our method transcends the limitations of textual matching and provides a scalable, automated alternative to labor-intensive manual inspection. This ensures that the agent’s final spatial products not only represent correct computational results but also adhere to established cartographic conventions and professional standards.

In terms of visual multimodal verification, we employ a reference-based visual comparative strategy. To minimize the inherent subjectivity in VLM-based evaluations, we synthesize a contrastive image by concatenating the agent-generated output with the ground-truth map generated via the execution of the gold-standard tool-invocation trajectory. Acting as a judge model, the VLM receives both the original task description and the contrastive image. Through a meticulous comparison of their visual and spatial characteristics, it quantitatively assigns a score on a scale of 0 to 100. The evaluation focuses on two core dimensions: (1) Data and Spatial Accuracy, which verifies whether the morphology, spatial topological relationships, and quantitative statistical results of the geographical features within the map strictly align with the ground truth, thereby capturing potential deviations in underlying data processing; and (2) Cartographic Style Adherence, which assesses whether the visual rendering—such as color ramp distribution and layer stacking order—conforms to user intent and established cartographic conventions.

\subsection{Efficacy Metrics}
Regarding the trajectory execution efficiency dimension, to quantify the workflow redundancy and resource utilization efficacy, this study defines an efficiency metric, $Eff$, based on tool-invocation trajectories. This metric encompasses both macro and micro levels, serving to characterize the average optimality of task execution and the global resource utilization rate, respectively. For the $i$-th successfully executed task, the step efficiency, $Eff^{(i)}$, is defined as:

\begin{equation}
\label{eq1}
Eff^{(i)}=\frac{N_{gt}^{(i)}}{\max(N_{gt}^{(i)},N_{pred}^{(i)})}
\end{equation}

Building upon this, we calculate the global efficiency using the following formulas:

\begin{equation}
\label{eq1}
\overline{Eff}_{macro}=\frac{1}{M}\sum_{i=1}^{M}Eff^{(i)}
\end{equation}
\begin{equation}
\label{eq1}
\quad\overline{Eff}_{micro}=\frac{\sum_{i=1}^{M}N_{gt}^{(i)}}{\sum_{i=1}^{M}\max(N_{gt}^{(i)},N_{pred}^{(i)})}
\end{equation}

\noindent where ${N_{gt}^{(i)}}$ and ${N_{pred}^{(i)}}$ denote the ground-truth and predicted step counts for task $i$, respectively, and $M$ represents the total number of successfully completed tasks. $\overline{Eff}_{macro}$ reflects the agent's average performance in path planning for individual tasks, whereas $\overline{Eff}_{micro}$ measures the system's capacity for redundancy control when handling large-scale task sets from a global perspective. By strictly bounding this metric within the interval $[0,1]$, we achieve a scientific measurement of the agent's operational conciseness and resource utilization efficacy during complex geospatial analysis processes.

\section{A Novel Plan-and-React Architectures}
The effectiveness of autonomous geospatial agents hinges on how their underlying reasoning paradigms handle the dual challenge of long-chain logical consistency and real-time data uncertainty. To identify the optimal reasoning logic for professional GIS workflows, we systematically evaluate three representative agent frameworks—Base Agent, ReAct, and Plan-and-Solve—to pinpoint their fundamental limitations. Based on these findings, we introduce the Plan-and-React architecture, a novel design explicitly engineered to overcome the trade-off between strategic planning and execution flexibility.

The Base Agent serves as the fundamental control group for tool-use evaluation. This paradigm equips the model with standardized tool definitions (schemas) and natural language instructions, enabling it to perceive real-time execution feedback from the dynamic execution sandbox. However, it lacks explicit multi-step reasoning or internal error-recovery mechanisms, primarily testing the model’s direct tool-scheduling capabilities in zero-shot scenarios. Without an internal reasoning loop, the Base Agent is highly susceptible to parameter hallucinations and struggles to maintain logical continuity in complex workflows.

The ReAct paradigm \citep{yao_react_2023} improves upon the Base Agent by following a canonical "Thought-Action-Observation" loop. In this architecture, the agent performs local reasoning at each step and dynamically determines the next action based on environmental observations. While this approach prioritizes real-time responsiveness, it frequently suffers from reasoning drift in long-chain geospatial tasks. Without a global roadmap to anchor its decisions, the agent may lose sight of the ultimate objective during deep tool-invocation sequences, leading to redundant loops or divergent analytical paths that eventually exceed execution limits.

The Plan-and-Solve approach \citep{wang_plan-and-solve_2023} addresses the drift issue by emphasizing global task decomposition before execution. This paradigm requires the agent to first generate a static sequence of steps as a comprehensive roadmap for the entire task. While it excels at the logical deconstruction of intricate geospatial problems, it exhibits execution rigidity when encountering unforeseen data anomalies. In the GIS domain, where coordinate mismatches or topological errors are common, the rigid "plan-first, execute-later" logic of Plan-and-Solve fails to recover from runtime errors. Once an intermediate step fails, the agent lacks the mechanism to adjust its trajectory, rendering the remainder of the plan unusable.

\begin{figure}[htb]
    \centering
    % 在单栏中，使用 \textwidth 是最标准的做法
    \includegraphics[width=\textwidth]{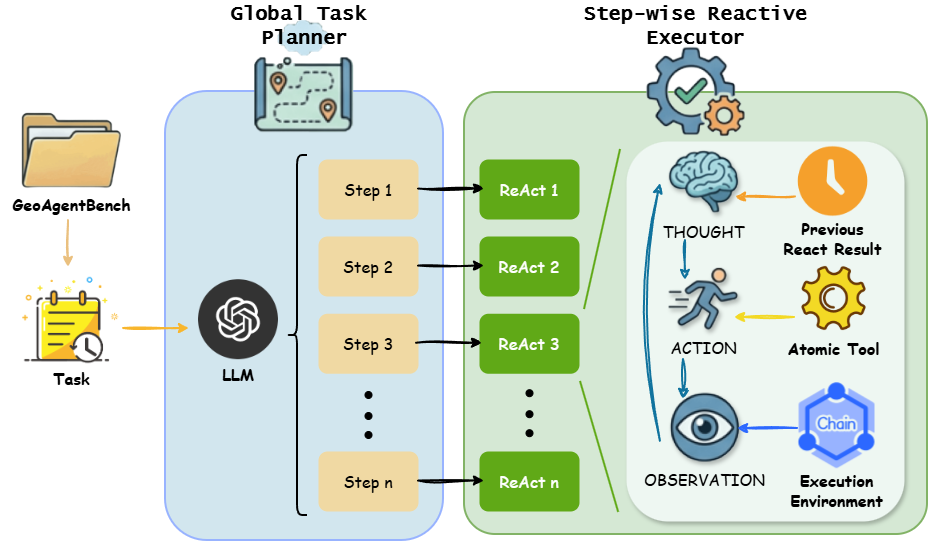} 
    \caption{The Plan-and-React baseline agent framework adopts a design that decouples global workflow orchestration from step-wise reactive execution. The Global Task Planner is responsible for decomposing abstract geospatial problems into logically self-consistent sequences of steps. Meanwhile, the Step-wise Reactive Executor implements tool invocation, dynamic parameter inference, and runtime error self-recovery through localized "Thought-Action-Observation" loops. This architecture mimics the cognitive paradigm of GIS experts, ensuring structural integrity while maintaining tactical flexibility in long-chain spatial analysis workflows.}
    \label{fig:baseline}
\end{figure}

Our proposed Plan-and-React architecture is explicitly designed to bridge these gaps. By recognizing that professional GIS operations are both strategically rigorous and tactically unpredictable, this framework structurally decouples the reasoning process into two synergistic components: a Global Task Planner and a Step-wise Reactive Executor

Instead of prematurely interacting with the environment, the agent first acts as a chief analyst through the Global Task Planner (as illustrated in Fig~\ref{fig:baseline}, 'Global Planning'). It comprehends the complex natural language instruction and formulates a holistic analytical blueprint prior to any tool invocation. This plan decomposes the overarching geographical problem into a logically self-consistent sequence of sub-tasks, establishing a deterministic analytical anchor. This blueprint effectively prevents the agent from falling into the infinite loops or reasoning drift that plague pure ReAct agents.

While the overarching plan provides strategic direction, our designed Step-wise Reactive Executor (see Fig~\ref{fig:baseline}, 'Step-wise Execution') handles the tactical uncertainties of multi-source spatial data. It processes each planned sub-task through a localized "Thought-Action-Observation" loop within the dynamic sandbox. The core innovation of this design lies in its constrained flexibility. If a tool invocation fails—for instance, due to a "Topology Error"—the executor does not abandon the global objective. Instead, it utilizes the execution feedback to autonomously diagnose the localized issue (e.g., repairing self-intersecting geometries) and retries the execution within the confines of the current step.

By seamlessly integrating global structural guidance with localized, feedback-driven error recovery, our Plan-and-React framework achieves an optimal balance between logical integrity and operational flexibility. It empowers the agent to maintain a clear analytical goal while dynamically adapting to the complexities of multi-source spatial data, thereby establishing a robust new standard architecture for the next generation of autonomous GeoAI systems.

\section{Experiments}
\subsection{Experiment Setup}
To comprehensively assess the performance boundaries of different agent architectures, this study selects a diverse array of models as experimental subjects, including mainstream open-source models such as Qwen2.5-7B-Instruct \citep{qwen_qwen25_2025}, Llama-3.1-8B-Instruct \citep{grattafiori_llama_2024}, and DeepSeek-V3 \citep{deepseek-ai_deepseek-v3_2025}, alongside high-performance closed-source models including GPT-4o \citep{openai_gpt-4o_2024}, GPT-4o-mini, Gemini-2.5-Flash \citep{comanici_gemini_2025}, and Claude Sonnet 4.6. Furthermore, to investigate the impact of different reasoning logics on geospatial analytical efficacy, each model is evaluated across four representative interaction paradigms: (1) Base Agent, (2) ReAct, (3) Plan-and-Solve, and (4) the Plan-and-React. All models were queried using default API configurations, and a standardized system prompt was employed to maintain consistency throughout the benchmarking process.

During experimental execution, the closed-loop GIS engine was configured with a maximum of 30 steps per task to prevent hallucination-induced redundant loops. Based on the stress testing of ground-truth execution times (with a peak duration of \textless300s), we rigorously established a 360-second execution timeout threshold for each individual tool invocation. This threshold is designed to effectively filter out invalid, long-running loops triggered by parameter configuration errors, thereby ensuring that the evaluation focuses on an agent's efficacy in orchestrating spatial analysis workflows. Through real-time monitoring and forced termination of task execution, we have established a reproducible and observable benchmark environment, providing a unified observational platform for in-depth analysis of agent success rates and logical reasoning biases across distinct spatial analysis tasks.

In the multimodal evaluation phase, we utilized GPT-4o as the judge model. Each "reference-prediction" image pair was subjected to independent repeated evaluations (n=3), with results ultimately presented as "mean $\pm$ standard deviation." Integrating ground-truth comparisons via visual inspection overcomes the limitations of traditional textual matching in assessing cartographic quality. Furthermore, the statistical method of repeated independent evaluations effectively mitigates the stochastic volatility inherent in model generation, quantitatively characterizing the robustness of the generated outputs in terms of cartographic conventions and spatial information representation. This ensures that the final evaluation metrics objectively reflect the comprehensive end-to-end spatial analysis performance of the agents.

\subsection{Result Under the Base Agent Paradigm}
According to the experimental results in Table~\ref{tab:base_agent}, the performance of various models under the Base Agent paradigm exhibits significant tiered characteristics. Closed-source frontier models, specifically Gemini-2.5-Flash and Claude Sonnet 4.6, maintain a leading position across all accuracy metrics. Among them, Gemini-2.5-Flash demonstrates the strongest performance in tool retrieval (TAO-F1: 82.48\%) and parameter execution (PEA: 43.02\%), while Claude Sonnet 4.6 excels in toolchain exact match (TEM: 53.01\%) and visual evaluation (66.57\%). The downward trend in metrics from TAO to TIO and then to TEM reveals the immense challenge of maintaining structural integrity within long-chain geospatial workflows. Notably, regarding the PEA metric, even the top-performing models fail to surpass the 45\% threshold. This underscores a universal deficiency in the models' implicit reasoning capabilities for professional GIS parameters—such as coordinate system encoding and spatial thresholds—when feedback mechanisms are absent.Furthermore, GPT-4o demonstrats the highest execution efficiency ($Eff$ \textgreater  97\%), whereas lightweight open-source models (e.g., Qwen2.5-7B and Llama-3.1-8B) exhibit a generational gap compared to top-tier models in terms of logical orchestration and visual output quality. These results suggest that while mainstream LLMs possess fundamental GIS tool-calling capabilities, simple zero-shot scheduling is insufficient to handle the rigorous logical dependencies and dynamic error-correction requirements inherent in geospatial analysis. This provides empirical support for the subsequent introduction of more sophisticated agent interaction paradigms.

\begin{table}[htb]
    \centering
    \caption{Performance evaluation of LLMs under the base agent paradigm.}
    \label{tab:base_agent}
    \begin{threeparttable}
    % 定义居中对齐的弹性列
    \newcolumntype{C}{>{\centering\arraybackslash}X} 
    
    \footnotesize % 10列大表必须使用 footnotesize 才能在单栏放下
    \setlength{\tabcolsep}{2pt} % 极限压缩列间距
    \renewcommand{\arraystretch}{1.3}

    % 1个左对齐列 + 9个居中弹性列 = 10列
    \begin{tabularx}{\textwidth}{l *{9}{C}}
        \toprule
        \multirow{2}{*}{Model} & \multicolumn{3}{c}{TAO} & \multirow{2}{*}{TIO} & \multirow{2}{*}{TEM} & \multirow{2}{*}{PEA} & \multirow{2}{*}{VLM\tnote{1}} & \multirow{2}{*}{$\overline{Eff}_{ma}$} & \multirow{2}{*}{$\overline{Eff}_{mi}$} \\
        \cmidrule(lr){2-4}
        & R & P & F1 & & & & & & \\
        \midrule
        Qwen2.5-7B & 17.94 & 34.91 & 22.01 & 13.56 & 13.56 & 12.42 & 8.1\tiny{$\pm$0.6} & \underline{95.6} & \underline{92.8} \\
        Llama-3.1-8B & 49.91 & 66.75 & 52.12 & 38.35 & 29.58 & 24.83 & 16.8\tiny{$\pm$1.3} & 72.2 & 56.9 \\
        GPT-4o-mini & 68.51 & 71.57 & 68.27 & 51.63 & 29.74 & 35.39 & 45.8\tiny{$\pm$1.4} & 85.5 & 81.1 \\
        GPT-4o & 67.04 & 72.98 & 68.95 & 51.51 & 45.80 & 32.54 & 51.0\tiny{$\pm$0.4} & \textbf{98.1} & \textbf{97.4} \\
        DeepSeek-V3 & \underline{82.20} & \underline{81.79} & \underline{79.78} & 62.55 & 50.80 & \underline{40.41} & 63.2\tiny{$\pm$1.2} & 92.0 & 86.6 \\
        Claude 3.5 & 79.17 & 72.49 & 74.70 & \underline{64.11} & \textbf{53.01} & 39.84 & \textbf{66.6}\tiny{$\pm$2.0} & 87.1 & 81.3 \\
        Gemini-1.5 & \textbf{82.60} & \textbf{85.89} & \textbf{82.48} & \textbf{66.16} & \underline{52.87} & \textbf{43.02} & \underline{66.3}\tiny{$\pm$0.8} & 89.8 & 86.6 \\
        \bottomrule
    \end{tabularx}

    \begin{tablenotes}
        \footnotesize
        \item[1] VLM-as-judge score; $\overline{Eff}_{ma}$ and $\overline{Eff}_{mi}$ denote macro and micro efficiency.
    \end{tablenotes}
    \end{threeparttable}
\end{table}

\subsection{Result Under the ReAct Paradigm}
Table~\ref{tab:react} presents the performance of various models under the ReAct (Thought-Action-Observation) paradigm. Compared to the Base Agent mode in Table~\ref{tab:base_agent}, the dynamic feedback mechanism of ReAct triggers a significant performance leap across all models. Claude Sonnet 4.6 exhibits the most superior logical orchestration and self-correction capabilities under this paradigm, with its tool retrieval (TAO-F1: 83.59\%), sequence consistency (TIO: 69.73\%), and parameter execution accuracy (PEA: 54.15\%) all reach peak levels, while its VLM visual score substantially improvs to 78.11\%. Notably, the ReAct paradigm offers the most pronounced boost to parameter execution (PEA), with leading models generally achieving gains exceeding 10\%. This provides strong empirical evidence that real-time execution feedback can effectively guide models in correcting parameter offsets during GIS tool invocation. As a representative of open-source models, DeepSeek-V3 performs strongly in visual evaluation (65.06\%) and macro-efficiency ($Eff_{macro}$: 89.95\%), demonstrating the potential to rival top-tier closed-source models. However, due to the iterative trial-and-error and retry processes inherent in ReAct, the execution efficiency ($Eff$) of the models declines relative to the Base mode, reflecting the computational overhead of redundant steps incurred to improve task success rates. In summary, the ReAct paradigm, by establishing a closed-loop feedback system, significantly bridges the gap between general-purpose LLMs and professional geospatial computing requirements, exhibiting enhanced robustness especially in complex tasks involving multi-step geometric transformations and spatial correlations.

\begin{table}[htb]
    \centering
    \caption{Performance evaluation of LLMs under the ReAct paradigm.}
    \label{tab:react}
    \begin{threeparttable}
    % 定义居中对齐的弹性列
    \newcolumntype{C}{>{\centering\arraybackslash}X} 
    
    \footnotesize % 10列大表在GSIS单栏中必须使用 footnotesize
    \setlength{\tabcolsep}{2pt} % 极限压缩列间距以适应单栏
    \renewcommand{\arraystretch}{1.3}

    \begin{tabularx}{\textwidth}{l *{9}{C}}
        \toprule
        \multirow{2}{*}{Model} & \multicolumn{3}{c}{TAO} & \multirow{2}{*}{TIO} & \multirow{2}{*}{TEM} & \multirow{2}{*}{PEA} & \multirow{2}{*}{VLM\tnote{1}} & \multirow{2}{*}{$\overline{Eff}_{ma}$} & \multirow{2}{*}{$\overline{Eff}_{mi}$} \\
        \cmidrule(lr){2-4}
        & R & P & F1 & & & & & & \\
        \midrule
        Qwen2.5-7B & 26.65 & 38.05 & 28.52 & 21.03 & 16.31 & 15.28 & 13.6\tiny{$\pm$0.9} & 85.8 & 66.2 \\
        Llama-3.1-8B & 42.53 & 64.16 & 47.64 & 35.41 & 30.86 & 25.85 & 7.7\tiny{$\pm$0.2} & 72.2 & 65.2 \\
        GPT-4o-mini & 67.15 & 67.54 & 65.71 & 46.02 & 31.08 & 32.70 & 36.9\tiny{$\pm$1.0} & 84.0 & 72.4 \\
        GPT-4o & \underline{80.37} & \textbf{85.81} & \underline{80.65} & 61.65 & 47.33 & 43.71 & 55.6\tiny{$\pm$0.8} & 85.4 & \textbf{82.6} \\
        DeepSeek-V3 & 78.58 & 81.74 & 78.16 & 59.63 & \underline{51.36} & \underline{46.74} & \underline{65.1}\tiny{$\pm$0.9} & \textbf{90.0} & \underline{80.8} \\
        Claude 3.5 & \textbf{88.25} & 81.49 & \textbf{83.59} & \textbf{69.73} & \textbf{55.00} & \textbf{54.15} & \textbf{78.1}\tiny{$\pm$3.5} & 82.4 & 75.0 \\
        Gemini-1.5 & 80.09 & 80.95 & 78.55 & \underline{67.69} & 48.89 & 45.51 & 64.3\tiny{$\pm$0.2} & \underline{86.2} & 78.1 \\
        \bottomrule
    \end{tabularx}

    \begin{tablenotes}
        \footnotesize
        \item[1] VLM-as-judge score; $\overline{Eff}_{ma}$ and $\overline{Eff}_{mi}$ denote macro and micro efficiency.
    \end{tablenotes}
    \end{threeparttable}
\end{table}

\subsection{Result Under the Plan-and-Solve Paradigm}
Table~\ref{tab:plan_and_solve} presents the performance of various models under the Plan-and-Solve paradigm, with its most striking feature being the drastic contrast between exceptionally high execution efficiency and extremely low output quality. Under this paradigm, due to the rigid "plan-first, execute-later" logic and the absence of runtime dynamic adjustment mechanisms, the execution efficiency ($Eff_{macro}$ and $Eff_{micro}$) of nearly all models reaches the theoretical limit of 100\%, indicating strict adherence to preset steps without any redundant attempts. However, this lack of feedback leads to catastrophic final results: VLM visual assessment scores collapsed across the board, with all models scoring below 4.0—far lower than the levels achieved under the Base Agent and ReAct paradigms. Although Gemini-2.5-Flash and GPT-4o still maintain a certain standard in tool identification (TAO-F1 \textgreater 83\%) and parameter execution (PEA), the high sensitivity of geospatial workflows to environmental states (such as file path dependencies or coordinate system transformations) means that any minor planning deviation results in total task failure during the execution phase due to the lack of fault-tolerance and recovery capabilities. This comparison forcefully demonstrates that in complex GIS scenarios, a linear execution mode—relying solely on macro-planning while lacking micro-level dynamic feedback—is entirely insufficient to meet the requirements of autonomous spatial analysis.

\begin{table}[htb]
    \centering
    \caption{Performance evaluation of LLMs under the Plan-and-Solve paradigm.}
    \label{tab:plan_and_solve}
    \begin{threeparttable}
    % 定义居中对齐的弹性列（如果之前定义过，可省略此行）
    \newcolumntype{C}{>{\centering\arraybackslash}X} 
    
    \footnotesize % 10列大表必须使用 footnotesize
    \setlength{\tabcolsep}{2pt} % 极致压缩列间距以平摊到单栏宽度
    \renewcommand{\arraystretch}{1.3}

    \begin{tabularx}{\textwidth}{l *{9}{C}}
        \toprule
        \multirow{2}{*}{Model} & \multicolumn{3}{c}{TAO} & \multirow{2}{*}{TIO} & \multirow{2}{*}{TEM} & \multirow{2}{*}{PEA} & \multirow{2}{*}{VLM\tnote{1}} & \multirow{2}{*}{$\overline{Eff}_{ma}$} & \multirow{2}{*}{$\overline{Eff}_{mi}$} \\
        \cmidrule(lr){2-4}
        & R & P & F1 & & & & & & \\
        \midrule
        Qwen2.5-7B & 29.53 & 35.79 & 31.27 & 25.01 & 21.64 & 8.24 & 1.7\tiny{$\pm$0.2} & 100 & 100 \\
        Llama-3.1-8B & 49.03 & 47.67 & 47.20 & 37.23 & 28.67 & 24.99 & 1.6\tiny{$\pm$0.2} & 100 & 100 \\
        GPT-4o-mini & 68.28 & 75.31 & 70.75 & 49.50 & 46.23 & 30.74 & \textbf{3.9}\tiny{$\pm$0.9} & 100 & 100 \\
        GPT-4o & \underline{81.58} & \underline{87.18} & \underline{83.63} & \underline{61.56} & \underline{58.18} & \underline{36.95} & 3.7\tiny{$\pm$0.2} & 100 & 100 \\
        DeepSeek-V3 & 76.22 & 79.43 & 76.72 & 60.01 & 51.99 & 37.82 & \underline{3.8}\tiny{$\pm$0} & 99.0 & 97.9 \\
        Claude 3.5 & 58.01 & 57.51 & 57.50 & 48.97 & 44.82 & 25.48 & 1.9\tiny{$\pm$0} & 100 & 100 \\
        Gemini-1.5 & \textbf{85.01} & \textbf{87.69} & \textbf{85.18} & \textbf{67.31} & \textbf{60.97} & \textbf{39.86} & 3.7\tiny{$\pm$0.1} & 100 & 100 \\
        \bottomrule
    \end{tabularx}

    \begin{tablenotes}
        \footnotesize
        \item[1] VLM-as-judge score; $\overline{Eff}_{ma}$ and $\overline{Eff}_{mi}$ denote macro and micro efficiency.
    \end{tablenotes}
    \end{threeparttable}
\end{table}

\subsection{Result Under the Plan-and-React Paradigm}
Table~\ref{tab:plan_and_react} presents the experimental results under the Plan-and-React framework, which achieves the optimal balance between logical rigor and execution success rate across all models. Claude Sonnet 4.6 attained the best overall performance within this framework, with its tool retrieval (TAO-F1: 84.94\%) and tool interaction order (TIO: 73.02\%) metrics both reaching their peaks, while its VLM visual score rose to 79.03\%, significantly outperforming any single paradigm. DeepSeek-V3 also delivered an excellent performance, demonstrating strong robustness in parameter execution accuracy (PEA: 47.34\%) and visual quality. The experiments prove that the Plan-and-React mode effectively reduces the blind trial-and-error inherent in the pure ReAct mode through global planning presets (as evidenced by the rebound of the $Eff$ metric compared to ReAct), while simultaneously overcoming the rigidity flaws of the Plan-and-Solve mode in geospatial environments via local reactive corrections. This synergistic effect not only substantially enhances the end-to-end success rate of agents in handling complex multi-step GIS workflow tasks but also establishes a solid technical benchmark for developing autonomous agents with professional-grade geospatial logical reasoning capabilities.

\begin{table}[htb]
    \centering
    \caption{Performance evaluation of LLMs under the Plan-and-React framework.}
    \label{tab:plan_and_react}
    \begin{threeparttable}
    % 定义居中对齐的弹性列（如果文档中已定义过，此处可不重复定义）
    \newcolumntype{C}{>{\centering\arraybackslash}X} 
    
    \footnotesize % 10列大表在GSIS单栏中必须使用 footnotesize
    \setlength{\tabcolsep}{2pt} % 极限压缩列间距以填满页面宽度
    \renewcommand{\arraystretch}{1.3}

    \begin{tabularx}{\textwidth}{l *{9}{C}}
        \toprule
        \multirow{2}{*}{Model} & \multicolumn{3}{c}{TAO} & \multirow{2}{*}{TIO} & \multirow{2}{*}{TEM} & \multirow{2}{*}{PEA} & \multirow{2}{*}{VLM\tnote{1}} & \multirow{2}{*}{$\overline{Eff}_{ma}$} & \multirow{2}{*}{$\overline{Eff}_{mi}$} \\
        \cmidrule(lr){2-4}
        & R & P & F1 & & & & & & \\
        \midrule
        Qwen2.5-7B & 41.86 & 54.03 & 44.79 & 26.49 & 16.14 & 16.90 & 20.1\tiny{$\pm$0.9} & \underline{86.9} & 62.8 \\
        Llama-3.1-8B & 44.26 & 59.34 & 47.11 & 34.07 & 18.93 & 21.59 & 10.1\tiny{$\pm$0.5} & 68.2 & 42.4 \\
        GPT-4o-mini & 74.08 & 71.00 & 70.95 & 57.80 & 41.10 & 36.15 & 43.4\tiny{$\pm$1.7} & 82.6 & 72.2 \\
        GPT-4o & 79.40 & \underline{83.70} & 79.55 & 61.38 & 38.58 & 40.69 & 59.1\tiny{$\pm$1.1} & 81.4 & \underline{79.3} \\
        DeepSeek-V3 & \underline{83.79} & 82.20 & \underline{80.64} & \underline{64.48} & \textbf{51.16} & \textbf{47.34} & \underline{68.5}\tiny{$\pm$1.5} & \textbf{89.5} & \textbf{86.2} \\
        Claude 3.5 & \textbf{89.00} & \textbf{82.95} & \textbf{84.94} & \textbf{73.02} & \underline{45.50} & \underline{46.11} & \textbf{79.0}\tiny{$\pm$1.5} & 79.5 & 73.3 \\
        Gemini-1.5 & 75.99 & 77.08 & 75.11 & 58.23 & 30.52 & 43.19 & 52.9\tiny{$\pm$0.2} & 79.6 & 77.5 \\
        \bottomrule
    \end{tabularx}

    \begin{tablenotes}
        \footnotesize
        \item[1] VLM-as-judge score; $\overline{Eff}_{ma}$ and $\overline{Eff}_{mi}$ denote macro and micro efficiency.
    \end{tablenotes}
    \end{threeparttable}
\end{table}

In summary, our extensive experiments across the four representative paradigms—Base Agent, ReAct \citep{yao_react_2023}, Plan-and-Solve \citep{wang_plan-and-solve_2023}, and the proposed Plan-and-React—reveal significant capability boundaries in autonomous spatial analysis. While Base Agents demonstrate basic tool-calling abilities, they struggle with the strict logical dependencies and long-chain reasoning inherent in professional GIS workflows. The ReAct paradigm improves runtime error recovery through its local "thought-action-observation" loops, yet it often suffers from reasoning drift or redundant loops when dealing with complex global objectives. Conversely, the Plan-and-Solve approach excels at macro-level task decomposition but exhibits limited flexibility when encountering parameter configuration errors or environmental anomalies due to its static execution nature.

The designed Plan-and-React framework achieves superior performance across almost all evaluation metrics, particularly in Parameter Execution Accuracy (PEA) and VLM-based end-to-end verification. By decoupling macro-level blueprint planning from micro-level reactive execution, this paradigm closely mimics the cognitive process of human GIS experts. It maintains a clear analytical goal while flexibly responding to data uncertainties and implicit parameter requirements. These results underscore that the synergy between global guidance and local feedback-driven correction is essential for navigating the inherent complexities of real-world geospatial analysis, establishing Plan-and-React as a robust baseline for the next generation of autonomous GeoAI systems.

\section{Conclusion}
In this study, we have presented GeoAgentBench (GABench), a pioneering dynamic and interactive evaluation benchmark specifically engineered for tool-augmented agents in the domain of Geographic Information Systems (GIS). By transcending the limitations of traditional static text and code-matching paradigms, GABench establishes a rigorous execution-based framework that integrates a professional-grade sandbox with over a hundred atomic GIS tools and a diverse array of complex, multi-step spatial analysis tasks. Our introduction of a multimodal evaluation mechanism leveraging Vision-Language Models (VLMs) further ensures that the performance of spatial agents is assessed not only on logical orchestration but also on the definitive accuracy and cartographic quality of the final spatial outputs. The experimental results across a spectrum of state-of-the-art Large Language Models reveal that while current foundation models exhibit remarkable potential for high-level task decomposition, significant challenges remain in implicit parameter inference and robust error recovery within complex geospatial workflows. Our findings emphasize that the synergy between global planning and local reactive debugging, as embodied in the Plan-and-React framework, is essential for navigating the inherent uncertainties and strict logical dependencies of real-world GIS operations. As the field moves toward the realization of truly Autonomous GIS, GABench provides the necessary scientific foundation and standardized metric system to guide the development of next-generation GeoAI systems. Future work will focus on expanding this benchmark to include more sophisticated spatiotemporal modeling and multi-agent collaborative workflows, further bridging the gap between general artificial intelligence and specialized geographic expertise to democratize and automate complex spatial problem-solving.

\bibliographystyle{abbrvnat} % abbrvnat 在作者超过 3 个时会自动压缩为 et al.
\bibliography{references}  %%% Uncomment this line and comment out the ``thebibliography'' section below to use the external .bib file (using bibtex) .

\end{document}